# Signs of the Past, Patterns of the Present: On the Automatic Classification of Old Babylonian Cuneiform Signs

ELI VERWIMP*, KU Leuven, Belgium
GUSTAV RYBERG SMIDT*, Ghent University, Belgium
HENDRIK HAMEEUW, KU Leuven, Belgium
KATRIEN DE GRAEF, Ghent University, Belgium

The work in this paper describes the training and evaluation of machine learning (ML) techniques for the classification of cuneiform signs. There is a lot of variability in cuneiform signs, depending on where they come from, for what and by whom they were written, but also how they were digitized. This variability makes it unlikely that an ML model trained on one dataset will perform successfully on another dataset. This contribution studies how such differences impact that performance. Based on our results and insights, we aim to influence future data acquisition standards and provide a solid foundation for future cuneiform sign classification tasks. The ML model has been trained and tested on handwritten Old Babylonian (c. 2000-1600 B.C.E.) documentary texts inscribed on clay tablets originating from three Mesopotamian cities (Nippur, Dūr-Abiešuḫ and Sippar). The presented and analysed model is ResNet50, which achieves a top-1 score of 87.1% and a top-5 score of 96.5% for signs with at least 20 instances. As these automatic classification results are the first on Old Babylonian texts, there are currently no comparable results.

CCS Concepts: • **Applied computing** → **Digital libraries and archives**; **Language translation**.

Additional Key Words and Phrases: Cuneiform, OCR, HTR, Old Babylonian Period, Sign Classification

## 1 Introduction

Automatic reading of cuneiform clay tablets from images is a difficult task and the handwriting characteristic of texts from the Old Babylonian Period (c. 2000-1600 B.C.E.) adds to the complexity. To let a machine read these texts, one needs to first localise the signs and subsequently classify them correctly. In this work, we present an analysis of how machine learning tools can automate the classification step, assuming either human or machine has first completed the localisation step. The analysis presented here uses a machine learning model trained on Old Babylonian documentary tablets.

The cuneiform script is mainly impressed onto clay tablets, thereby making it a 3D script. Shadows cast by incoming light heavily influence the legibility of the texts, both for humans and machines. When using photographs or scans of tablets with a single and often non-slanting light source, signs are less legible. To obtain optimal machine solutions, it is crucial to examine how a tablet needs to be visualised. We tested various lighting angles, normal maps and other depth visualisations to better understand which visualisations yield superior results. Our source corpus, which consists of 2D+ image data [11], is ideal for this purpose.

Furthermore, cuneiform tablets in the OB Period are highly diverse (see Section 3.1.2). We therefore study the transferability of models trained on subgroups of our corpus, specifically on place of origin, and the effects of a fine-tuning step. Using the data representations created by the classification model as an input for TSNE [30] visualisations, variations between the subgroups can be examined. Such understanding can inform us on the difficulties and the potentialities of generalizing to unseen subgroups of tablets.

---

*Both authors contributed equally to this research.
Our code base is publicly available at: https://github.com/VerwimpEli/signs-of-the-past

Authors' Contact Information: Eli Verwimp, eli.verwimp@kuleuven.be, KU Leuven, Leuven, Belgium; Gustav Ryberg Smidt, GustavRyberg.Smidt@UGent.be, Ghent University, Ghent, Belgium; Hendrik Hameeuw, KU Leuven, Leuven, Belgium; Katrien De Graef, Ghent University, Ghent, Belgium.

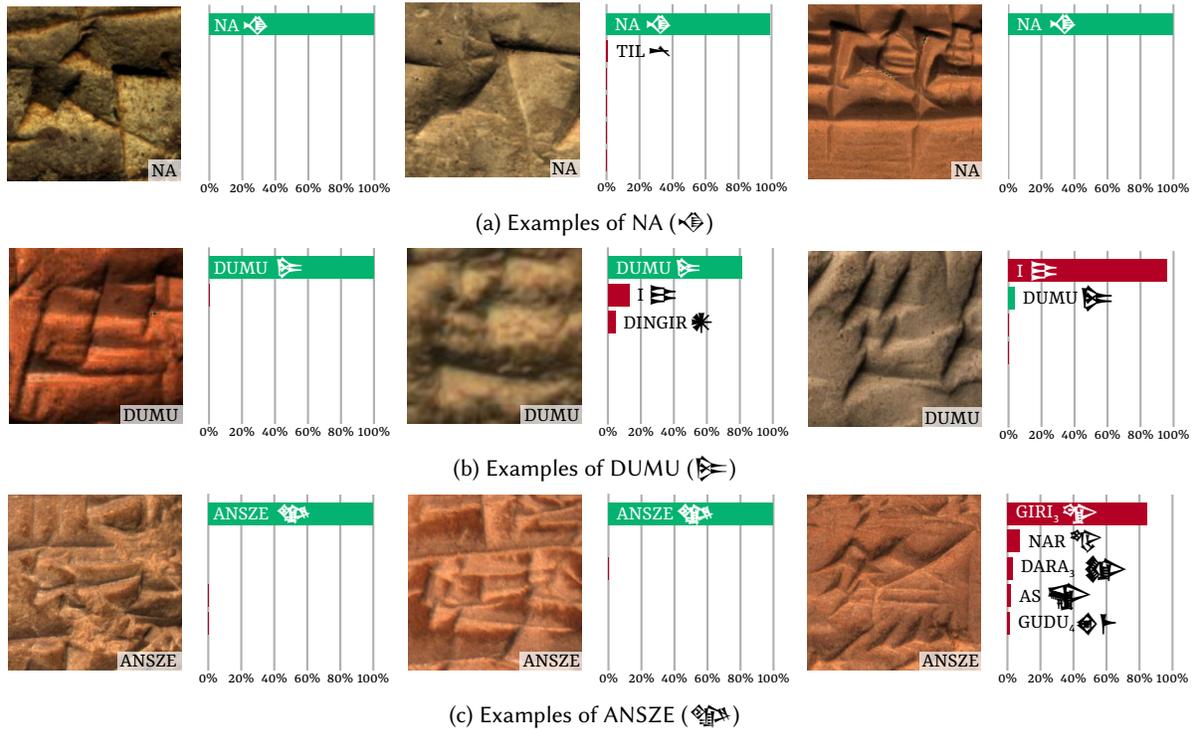

Fig. 1. Examples of three sign categories that are included in the data used in this paper, together with the predictions of a classification model trained as described in Section 3.3 for those signs. Only predictions with a confidence score higher than 0.5% are shown. The green bar indicates the correct sign categories, the red bars are wrong predictions. The predictions shown here are from a model trained on 'ColorA' visualizations, see Section 3.2.1. The examples shown here are from the accompanying 'ColorA' test set.

Figure 1 shows nine examples of the data that we have used in this paper. These images are visualised with light coming from above the tablet. Next to each image, the top-5 predictions of the trained model, as explained in Section 3.3, for these example signs are shown. In this paper, we will qualitatively analyse some of the results that are visible in this figure. We examine how variations in signs categories are dealt with by a trained ML model, how different visualizations influence the results and why in some cases, signs are still classified wrongly.

## 2 Related Work

The application of OCR techniques in humanities research has had a considerable impact over the last few years [21], for instance, the character error rates for printed medieval books in the Latin script can in many cases be below 1% [24]. When applying OCR techniques on ancient languages there are several factors that complicate such efforts. Some languages have only very limited data available, various writing methods and materials are in use, and some are still undeciphered, see Sommerschield et al. [27] for an overview. Here, we focus on the cuneiform script, which has its own particular challenges when applying OCR techniques. In the next paragraphs and Section 3.1, we will elaborate on these specific challenges.

Traditionally, the automatic recognition of cuneiform signs has put emphasis on a two step approach. The individual wedges and their orientation were identified first, and then signs were classified based on the arrangement of the individual wedges, see Bogacz et al. [3] for a summary. Such approaches have the advantage of being applicable on unseen data, given it follows the same system. However, the cursive writing of the presented corpus makes these approaches less useful. Lately, advances in ML research have made it possible to operate directly on images of cuneiform tablets. Williams et al. [33] study the full pipeline by breaking automatic recognition into three tasks: line estimation, sign localization and sign classification. Cobanaglu et al. [7] perform a similar evaluation with a focus on the types of ML models used and a larger dataset.

There are many cuneiform tablets available that are not fit to use directly as training data. Most tablets only have transliterations associated with them, without bounding box annotations. Dencker et al. [9] use a weakly supervised algorithm to iteratively align the transliteration with signs on the tablets. In doing so, more annotated data can be created, which is especially relevant for rarer signs. Mikulinsky et al. [20] propose to generate more data for those rare signs by training a diffusion model on sign prototypes, which had been studied earlier using conditional GANs [4]. In this paper we have a dataset consisting of signs manually annotated with location and transliteration, so no data generation or automatic labelling is used here.

Cuneiform signs are 3D impressions. When using images as input of an ML model, the depth information of the impressions needs to be carefully transferred to the 2D plane of the image. Typically, directional light is used to cast shadows which reveal depth information. Rest et al. [23] and Stötzner et al. [28] both study how a model trained with light coming from one angle transfers to another, and how using different (virtual) light sources can make models more robust against changes in the angle of incoming light. When 3D models or normal maps are available, the local surface orientations of the tablet can be used to make wedge impressions more explicitly visible than with shadows and directional light, see e.g. [1, 12]. Here, we work with both photorealistic images and non-photorealistic depth visualizations, to better understand how to optimally visualize cuneiform tablets for AI tasks.

Cuneiform writing has been used in many different places (space) and periods (time). Many signs change appearance based on where and when they were written. This is most explicitly studied by Yugay et al. [36] who try to distinguish between Neo-Assyrian and Neo-Babylonian signs using a ML model. Other papers also work on Neo-Assyrian data [7, 9, 20], Elamite [7, 33] or Hittite [23, 25]. We use Old Babylonian data, which to the best of our knowledge has not yet been studied as a stand-alone corpus in a sign classification task. Where most research has focused on various methodological applications, we explicitly study the challenges relevant for Old Babylonian clay tablets, such as geographical variations, data acquisition types and the effects of handwriting.

## 3 Methodology

### 3.1 Corpus Information

*3.1.1 Cuneiform and Akkadian.* The cuneiform script is named after the wedge shape indentations that make up the basic elements of a sign (wedge in Latin is 'cuneus'). It was in use from c. 3400 B.C.E. until c. 100 C.E. and likely invented for administrative purposes by people speaking the Sumerian language in the urban area of Uruk. Thereafter its use spread over large parts of western Asia i.e., Fertile Crescent and adjacent areas.[1] The script uses a mixture of logograms i.e., signs with a semantic component, and syllables, which makes it a logo-syllabic script. Cuneiform also has classifiers that specify the semantic character of a word, and numerical signs. Signs can express multiple readings that can be discerned from context, these we will refer to as variations or variants, whereas each sign will be a class and given a class name. For instance, the sign we give the class name NI (▷) can have multiple reading variations such as NI, LI$_2$ and ZAL.

---

[1]For a recent exploration of the spread of cuneiform sources see [22].

Most cuneiform texts were written on clay, which after drying or firing become solid and hardy objects that can withstand the test of time. The tablets used in this paper are mostly rectangular and can have bulgy surfaces with the curvature on each side of the vertical axis, thereby making the left and right sides typically slanting towards the edges. Cuneiform was predominantly impressed into clay tablets with a reed stylus giving the script its 3D characteristics. The depth of the signs makes their reading depending on lighting angles and for practical reasons, it can be assumed that the ancient scribes would have positioned the tablets so the light would be angled from above.[2] During writing the clay tablets are wet or damp and as such they are very plastic, resulting in writing that can be very diverse. The choice of sign variations can change within one text and scribal traditions, that varied from city to city across the history of cuneiform writing. These aspects of cuneiform writing on clay tablets are the results of them being handwritten.

Across the history and the geographical spread of cuneiform in what is the modern-day Middle East, a multitude of languages were written in the cuneiform script e.g., the Indo-European language Hittite, the linguistic isolate Sumerian and the Semitic language Akkadian. Akkadian texts dominate the cuneiform corpus in numbers, followed by Sumerian [29]:38-9. The work of this research is focused on Akkadian and specifically Akkadian documentary texts[3] from the Old Babylonian (OB) Period (c. 2000-1600 B.C.E.) in what is today southern Iraq. Within this area, typically known as Babylonia and named after the historically significant city of Babylon, the dominant powers in the OB Period changed and rarely did one centre hold the power over the whole of Babylonia.

Surfaces of OB Akkadian clay tablets are typically read in the order of: front, bottom, back, top and left. The order can change as not all surfaces are inscribed on each tablet. When going from front to back, a tablet is flipped along its horizontal axis, differently to what we do with pages in books. The right surface contains spill-over lines from front and back, and rarely from the bottom and top. This means it is common to find signs rotated according to the expected reading direction; writing on the left edge is often rotated 90° to suit the writing surface better and writing from the back running over the right is considered 180° rotated compared to the writing from the front running over the right surface.

*3.1.2 Corpus variations.* Across all of its approximately 3500 years of history the cuneiform script and the languages written in it, changed considerably. Similarly, but to a lesser degree, do we see differences in the OB Akkadian documentary texts from both geographical and chronological perspectives. The often changing political landscape of the OB Period meant that scribal traditions of different cities could wax and wane in influence. With the political turmoil near the end of the reign of Samsu-iluna (1749-1712 B.C.E.) and the movement of people it resulted in [6]:372, scribal traditions and tendencies also changed. Of strong traditions, the Nippurean is noticeable, it has a heavy focus on written Sumerian in the city during the OB Period [26]:130. Interestingly, the unlocated city of Dūr-Abiešuḫ, that shows close historical links to Nippur, does not exhibit the same "Nippur-Sumerian" vocabulary [18]:6. The survival of spoken Sumerian in the OB Period is a contested topic,[4] but certainly the OB Akkadian scribes would read Akkadian even when writing Sumerian [26]:129,136, a phenomenon Gershevitch coined as alloglottography [10]. This means that a text written with more or less exclusively Sumerian signs would have been read and understood in Akkadian. A last variation to be mentioned is that of genre. Documentary texts cover many genres and some are very formulaic e.g., contracts, whereas others are less formulaic and can vary much between texts of the same genre e.g., personal letters or inheritance documents.

As a consequence, our corpus has little sign standardization. To be able to read cuneiform clay tablets it is necessary to have an intuition of the essential graphical elements of a cuneiform sign and to know the context when visual variations are not discernable. Some good examples of this aspect are shown by the qualities and shortcomings of our trained model. First, in Figure 1a, the often occurring NA (𒈾) has some overall features that makes it easy to discern, and the model is then able to handle the internal variations of NA. Whereas, the signs DUMU (𒌉, logogram for 'son') and I (𒉌, syllabic sign) are both rather similar looking signs. To discern between them is sometimes confusing for our model on purely graphical terms. As Figure 1b shows, the model can confuse exactly those two signs. However, as the functions of the two signs vary greatly, it is typically easy to distinguish between them based on context. In Figure 1c we see how variations of a sign, in this case ANSZE (𒀲), are easily discerned when they are somewhat similar (the two left examples). Yet a different, and in our corpus rarely attested, variation of the same sign (right example) does not have the correct sign class in its top five predictions given from our model.

---

[2]Deciphering cuneiform tablets in modern times is usually done with the lighting angle from the upper left corner.
[3]For the purpose of this paper we define the term *documentary texts* as texts documenting reality e.g., letters, contracts or accounts of workforce.
[4]For discussions and further references see [19, 34].

Many regions where cuneiform tablets were discovered in the past have long suffered from instability, largely as a result of Western Imperialism. Besides the immense human suffering it has led to, cultural heritage of the region has been taken for museums and private collections worldwide. For this paper, these conditions for the cuneiform cultural heritage offer two specific challenges, namely the matters of (in)accessibility and image quality. The former can be a matter of a cuneiform object being far away from the researcher who wishes to access it, a lack of a digital twin of the object or rights for publishing the object being limited by the collection. Where images can either be accessed or acquired from a collection, the question of image quality becomes relevant. The cuneiform corpus is vast and the larger collections have tens if not hundreds of thousands of tablets and tablet fragments to image and publish. To best read cuneiform tablets light must hit the tablet at an angle when imaging (see Section 3.1.1), but it can be labour intensive work to acquire images with a variation of lighting or 3D models. Common practices are therefore to produce flatbed scans of the tablets or photograph them in a static light setup i.e., tablets are rarely photographed with lighting producing sufficient shadows of the signs on them.

### 3.2 Data

*3.2.1 Collections and proveniences.* As training and test datasets in this paper, there are currently three datasets defined by both collection and place of origin: Nippur texts from the Frau Professor Hilprecht Collection of Babylonian Antiquities at the Friedrich-Schiller University in Jena (HS), Dūr-Abiešuḫ texts from the Department of Near Eastern Studies at Cornell University (CUNES) and Sippar texts from the Vorderasiatische Museum in Berlin (VAT). Additionally, a test only dataset is available consisting of texts from the Royal Museums of Art and History (RMAH) in Brussels, where 8 tablets from the city Marad have been selected to take the role of an unseen provenience to test the out-of-distribution generalization of the trained ML model. Nippur, Sippar and Marad have been geographically localized (see Figure 2a), but where Dūr-Abiešuḫ was, is still uncertain, the current best estimate is that it was in the close vicinity north of Nippur [2]:27. As mentioned above, Dūr-Abiešuḫ has relations to Nippur, it seems to have been a fortress built for inhabitants of Nippur when the First Dynasty of Babylon experienced upheaval that threatened the safety of the religiously important city of Nippur. Differently, is it for both Sippar and Marad, the former is a main centre itself, where the latter might have been more linked to the city Isin [8].[5] All texts have been acquired without archaeological information, but it has been possible based on palaeographic and prosopographic studies to ascertain the origin of the tablets.

Within this corpus with different geographical proveniences there are chronological variations that extend within the OB period. Since Dūr-Abiešuḫ functioned as a safe haven for Nippur citizens in the later part of the OB Period and that a lot of the scribal activity was moved from Nippur to Dūr-Abiešuḫ [2]:32, it is then logical that the chronological spread of tablets from Nippur predates tablets from Dūr-Abiešuḫ. The tablets from Sippar with dates are predominately from around the reign of Hammurabi (c. 1800-1750 B.C.E.), which is earlier than the Nippur/Dūr-Abiešuḫ split. Not all tablets and genres are provided with dates, that is why it is difficult to define their chronological spread precisely. We have therefore chosen to not test on chronological divisions of our corpus yet.

---

[5]In Smidt and Verwimp forthcoming, we present a methodology based on the ML model presented here that attempts at quantifying the linguistic similarities between different cities.

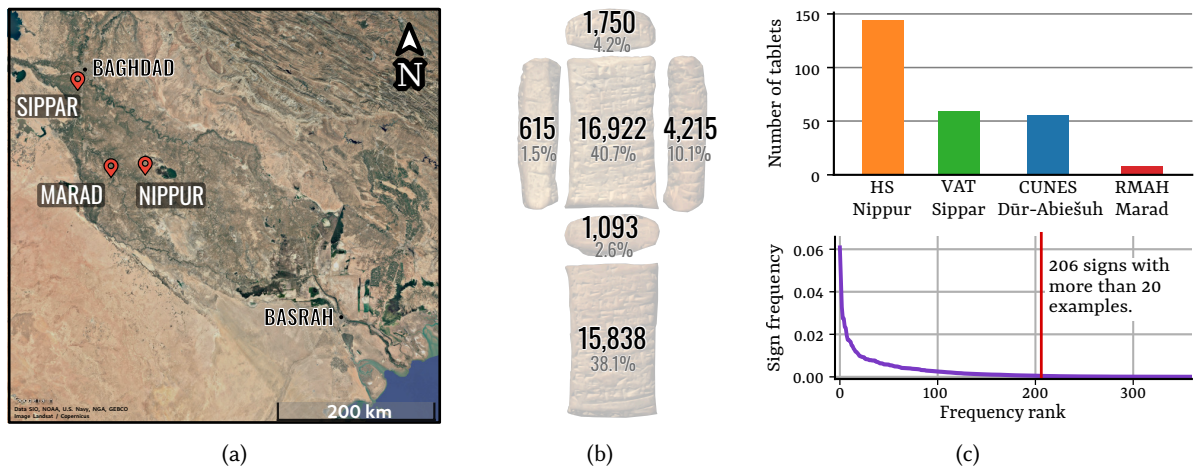

Fig. 2. Overview of the data that is used in this paper. Left (a): map from Google Earth of southern Iraq, with the locations of Sippar, Marad and Nippur, Dūr-Abiešuḫ is not located but presumed to be to the north and within a few kilometers of Nippur [2]:27. Middle (b): sign count per tablet side for all tablets from HS, VAT and CUNES. Right top (c): number of tablets per provenience. Right bottom (c): frequency of individual signs in the complete text corpus. 206 of the 360 signs appear at least 20 times and are used as training and test data (97.9% of all signs).

*3.2.2 Data acquisition.* The data for the classification task consists of visualisations and sign annotations. Visualisations acquisitions were made with several versions (various camera resolutions) of the White Light Portable Light Domes [32]. The acquisitions were done by different people, at different collections and at different times. We have chosen to name the datasets after the proveniences they are assumed to be from. Once processed with a photometric stereo algorithm, the resulting 2D+ models allow the interactive virtual relighting of photorealistic visualisations and non-photorealistic representations that reveal depth information of the tablet. As mentioned above, cuneiform clay tablets are handwritten texts with impressed signs; lighting angles and depth information are therefore important factors when reading cuneiform clay tablets. For the Nippur, Dūr-Abiešuḫ and Sippar data, visualisations were made of individual surfaces of the tablets that contained text. Based on their 2D+ models 12 different types of visualisations (2D raster images) were prepared for each surface, eight of them corresponding to different standardized lighting angles (ColorA-G), one with even lighting, mimicking flatbed scans (Color00), a normal map and two derivatives thereof, based on the curvature of the normal map (SketchA/B). The naming convention of the visualisations corresponds to Hameeuw et al.[11]:236. The Marad tablets from the RMAH were acquired with the same methodology, but we only evaluated their SketchB visualisation.

Images were annotated in the Cuneur software [14], where the tablets are displayed directly through International Image Interoperability Framework (IIIF) links of the JPG[6] image files.[7] Each sign on the tablet is manually enclosed in a polygonal figure by an annotator, and the figure is given a sign value according to the contextual based reading.[8] The data and scripts published with this paper can be found on our Zenodo/Github page (link).

---

[6][11] writes the file format is PNG, but that was changed to JPG throughout the project.
[7]Foket et al. forthcoming will account for the choice of IIIF as a framework for data sharing within the Cune-IIIF-orm project.
[8]Smidt forthcoming will present an evaluation of the annotation data.

The training and testing data for this study i.e., the groups Nippur, Dūr-Abiešuḫ and Sippar, consists of 258 tablets and the Marad testing data is 8 tablets (see Figure 2c top). There are 360 different classes for the classification task in the data, whereof 206 classes have at least 20 instances (see Figure 2c bottom). The largest corpus, from Nippur, is slightly larger than the Dūr-Abiešuḫ and Sippar sets together, the latter two are of similar size. Of the signs, 78.8% are annotated on the front or back (see Figure 2b). To discern between different sign classes we used the Nuolenna Unicode character to sign list [16].[9] It allowed us to compile a list of sign variations that shared a Unicode character and we then gave these characters names according to a common variation of the signs to better recognise them.

### 3.3 Model training

The data samples are cropped out of the full tablet images using their surrounding polygon. This polygon is transformed into a square bounding box, by taking the rectangular box at the extreme points of the polygon and extending the shorter sides to square it. Doing so often results in other signs becoming partly visible in this cropped box. Initial experiments revealed that those partial signs were useful training noise and led to trained models that made more robust predictions.

As classification network, the standard convolutional network ResNet50 [13] is used, which performed slightly better than the smaller ResNet18 and on par with ResNeXt50 [35]. This model is optimized with the AdamW optimizer for 30 epochs and batch size of 64 examples. The learning rate is set to the standard value of 0.001, weight decay is used with strength 1e−5 and a cosine learning rate scheduler shrinks the learning rate to a minimum value of 1e−5. For some experiments, models go through a final '*fine-tune*' training stage, where they are further trained with no augmentations and learning rate decaying from 0.0005 to 1e−7.

By default, data augmentations are used to make the training data more diverse. Every data point is first normalized and resized to 224 × 224 pixels. Then two geometric transformations are randomly applied with 50% probability: random rotation anywhere between 0° and 360°, and a random perspective transformation, which creates the illusion that a sign was photographed from a different angle. Simple colour augmentations did not improve initial results, so they are not used (see [23] for more complex colour augmentations). As mentioned in Section 3.2.1, cuneiform signs written on the sides of the tablets, other than the front and back, often appear rotated and since tablets are usually not flat, the perspectives of the data points vary. There are a few signs that may change semantics when rotated e.g., ASZ (←) and DISZ (↑), but when evaluating this aspect, it does not influence the effectiveness of the predictions as long as some parts of other signs are visible to determine the writing direction.

### 3.4 Experimental Details

The output of most ML models is an $n$-dimensional vector, which contains the likelihood for each of the $n$ possible categories. We report top-1 and top-5 accuracy, which are the percentage of test examples where the correct category is the most confident (top-1) or within the five most confident predictions (top-5). Each training experiment is repeated five times and reported as mean ± standard deviation. For every sign category, 80% of the data examples are used as training data and 20% as test data. We exclude categories with less than 20 examples, because with fewer examples learning becomes hard, and no reliable estimates of the performance for those categories can be made. To interpret some of the results, TSNE [30] plots are used to visualize the feature representation of a trained model i.e., the output of the second to last layer. For the ResNet50 model we use, the dimension of this vector space is 2048. While TSNE plots are a useful tool, they should be interpreted with care. In some cases, even randomly distributed data can appear clustered, and vice-versa (see e.g., [31]).

---

[9]The list is available on Github: https://github.com/tosaja/Nuolenna/blob/master/sign_list.txt.

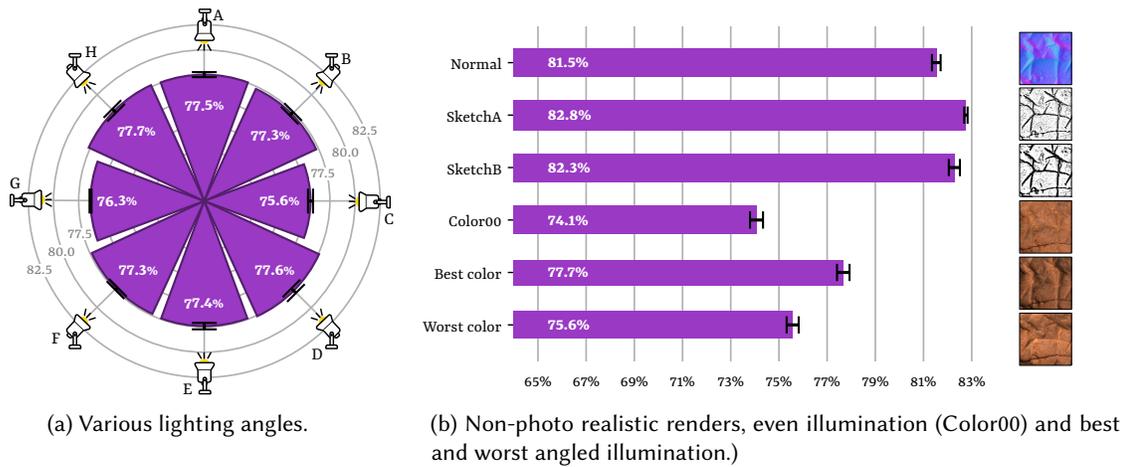

(a) Various lighting angles.

(b) Non-photo realistic renders, even illumination (Color00) and best and worst angled illumination.)

Fig. 3. Results for models trained on different visualisations of the tablets. Each result is the average top-1 accuracy, i.e. the percentage of correctly classified signs, on a dataset combining Nippur, Dūr-Abiešuḫ and Sippar signs, with a single visualization. The best colour on the right is 'ColorH', the worst is 'ColorC'. The images on the right show the different visualization of the same sign, in this case an instance of the sign 'NA' (𒈾).

## 4 Results and Discussion

This section first presents some overall results, then discusses the influence of the imaging technology on sign classification, followed by results on the availability of the data and the impact of writing variations across tablets from different proveniences.

### 4.1 Results

When training a base model with data from Nippur, Dūr-Abiešuḫ and Sippar as presented in Table 1, the top-1 accuracy has a maximum performance of 85.6% ± 0.2 on the Nippur set and a minimum of 77.2% ± 0.2 on Dūr-Abiešuḫ data, whereas for top-5 accuracy the range is 96.1% ± 0.1 on Nippur data to 91.9% ± 0.6 on Sippar data. The overall best score is our model fine-tuned for Nippur data and then tested on Nippur material gives 87.1% ± 0.3 (top-1) and 96.5% ± 0.1 (top-5). It is no surprise that the performance on Nippur data is consistently better since its training corpus is the largest.

### 4.2 Influences of Imaging Technology

Training on individual visualisations of the tablets allow us to study their effects on the legibility for the ML model of cuneiform signs. The signs are 3D impressions, and varying light angles are typically used to cast the right shadows to be able to read a sign. Alternatively, non-photorealistic renderings can be used to combine information from different lighting angles. In this discussion we only consider the influence of imaging technology on shadows for the legibility, since texture and colour[10] are generally irrelevant for the reading of cuneiform tablets.

When training on datasets of photorealistic visualisations limited to individual lighting angles, the lighting angle from upper left (ColorH) performs the best (77.7%), but the performances for all lighting angles, except for light coming directly from the right (ColorC) or left (ColorG), are within the margin of error (see Figure 3a).

---

[10]An experiment where the RGB-values of 'ColorA' are averaged to greyscale values did not lead to a change in performance.

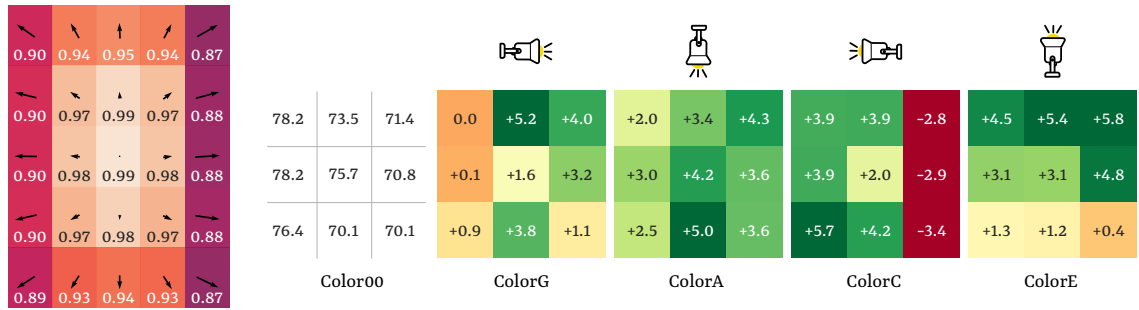

(a) Average normal vector of signs. The arrow indicate the $x$ and $y$ component, the number and colour the $z$ component.

(b) Absolute accuracy on a three by three grid for Color00, and the accuracies relative to those of Color00 for four different illumination angles.)

Fig. 4. The location of a sign on a tablet has a significant influence on the classification accuracy. To study this we divide every tablet side in a five by five (left) or three by three (right) grid and compute the average normal and accuracy for each of these sections.

Photorealistic 2D representations of 3D objects can not explicitly convey depth information, except by the shadows that are cast when light falls from a certain angle. When the angle is close to being parallel with the surface of the tablet, the entire sign can be covered in shadows. When the angle is closely aligned with the normal direction of the sign instead, no or only minimal shadows are cast. Due to the shape of the volume of a cuneiform tablet a single lighting condition will not reflect all signs in a similar manner. The average normals of a cuneiform tablet is pointing outwards, especially on the left and right edges as can be seen in Figure 4a. When light then comes from the left or right, only few shadows are cast on signs with a normal pointing towards the direction of the light, which explains the worse results for ColorG and ColorC. In Figure 4b we break down the classification accuracy on nine sections of the tablets with selected lighting angles compared to the even illumination (Color00). The results of Color00 thus serve as the baseline for the difficulty of classifying the signs within a given section. The relative differences are then largely the result of changing light directions. Figure 4b shows that when normals and light directions are aligned, the performance decreases. This effect is the strongest, where the tablets curve the most, i.e. the right side and to a lesser degree the left side.

Despite performance being almost equal (see Figure 3a), the partitions in Figure 4b reveal that models trained on data with different lighting angles, do not classify the same sign instances equally well. The shadows that are cast are different, hence the results are too. In capturing the images of the tablets, the White Light Portable Light Dome uses the shadows in all directions to create normal maps [32]. Figure 3b shows training on the normal maps is beneficial (81.5%), with a smaller additional improvement for 'SketchA/B' (82.8% and 82.3% respectively). The normal maps can contain confusing information. The $x$ and $y$ components of the normal maps are stored in the red and green channel respectively, which makes them not rotation invariant. However, the sketches are rotation invariant, which notably improves the results on the top, bottom, left and right sides of tablets, where signs are typically rotated[11].

---

[11]This is most extreme on the left side. The model trained on the normal maps classifies those signs correctly with 38.6% compared to 70.0% for SketchB.

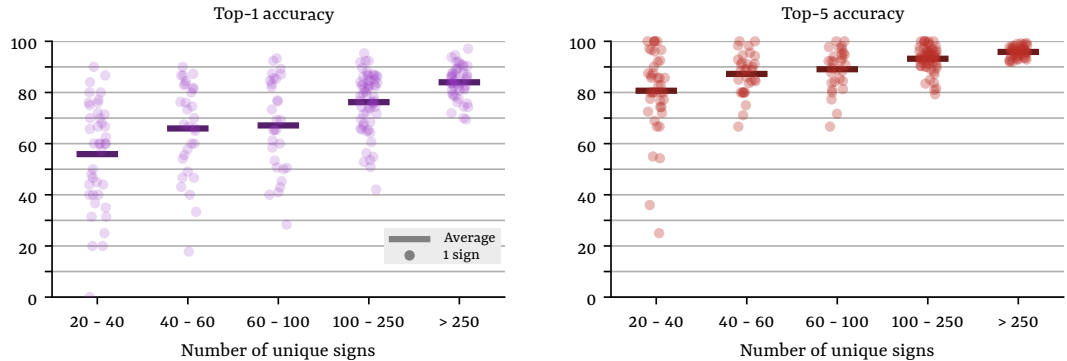

Fig. 5. Top-1 and top-5 accuracy of different sign categories for a model trained on the SketchB visualisations, grouped by the number of unique signs. Every dot indicates the accuracy of a different sign and the horizontal bars indicate the average for that group. In total there are 206 unique sign categories.

### 4.3 Influence of Data Availability

In general, machine learning models tend to perform better when more relevant and varied data becomes available. Nonetheless, we can highlight a few tendencies of the data that seem particularly noteworthy to consider when doing data acquisition and pre-preprocessing.

*4.3.1 Results by sign class.* While 206 sign classes with 20 instances or more are already included in the dataset, many more exist.[12] Figure 5 gives an indication of how much data needs to be collected and annotated to get reasonable performance. The average top-5 accuracy for the lowest category (20 to 40 signs) already reaches 80.7%, yet for a comparable top-1 performance (76.2%) between 100 and 250 attestations of each sign class are necessary. Random guessing performance for top-5 and top-1 accuracy with 206 classes is 2.42% and 0.48%, respectively.

There are many sign categories with fewer than 20 examples in our database. We did not include them here, as with so few signs, it is hard to provide a reliable estimate of the performance. The number of attestations per sign class can also vary considerably depending on how signs are counted. For instance, Sîn is a god whose name is typically written with three parts: the divine classifier DINGIR (✳), and the logograms EN (𒂗) and ZU (𒍪). EN is a sign with empty space in the upper left corner, wherein the star-looking sign DINGIR fits well, which is presumably why it is often written there in combination with EN. Counting the two signs individually would add a count for DINGIR and EN, but do to their combined state they could be assumed as one sign. Similarly, other combined signs could be considered one or more.[13]

*4.3.2 Qualitative analysis.* Ideally, enough training data ensures that a ML model ignores irrelevant background information, like image sharpness, or the surface of a tablet. Yet some variability in the data is inherent to cuneiform writing, where multiple variations of the same sign exist, and they can be written with different degrees of obliqueness. These differences have real implications. A model trained on only the Nippur signs classifies SZU (𒋗) from Nippur tablets with 75.0% accuracy, but SZU signs from Dūr-Abiešuḫ and Sippar with 47.9% and 56.7% accuracy, respectively. For UD (𒌓), those of Nippur tablets are classified correctly with 92.5% accuracy, while those of Dūr-Abiešuḫ and Sippar with only 22.2% and 14.09% accuracy. The variations of SZU are more similar across provenances, but they are still somewhat closely clustered, as indicated by the TSNE plots (Figure 6a). By a qualitative examination of the clusters, some variations akin to scholarly divisions are evident. The sign SZU is the sign for a 'hand' with a vertical wedge representing the palm and horizontal wedges representing fingers with the bottom finger being the thumb. The clusters of SZU cover variations of the size and distribution of fingers, one variation has a big thumb and a larger little finger with a variable number of smaller fingers in between the two (see Figure 6a, I). Another variation (Figure 6a, III) has the thumb being larger than the other fingers that are all of somewhat equal size. Other variations fall in between those two (Figure 6a, II) and some are what we would consider cursive or oblique, so typically written more slanted. Labat, who compiled one of the most thorough palaeographical sign lists of Akkadian cuneiform, has similar variations [17]:162. UD is the sun rising in a valley between two mountains turned 90° counter-clockwise. It mainly varies in the number of vertical wedges representing the mountains and valley, either one (Figure 6b, I&II) or two (Figure 6b, III). And for the signs with one vertical wedge, the cursiveness of the two angled wedges on the left representing the sun changes e.g., compare Figure 6b, I with Figure 6b, II. These variations are also in line with [17]:174. In the TSNE plot, the cluster most similar to Figure 6b, III is dominated by Nippur texts with some interspersed attestations from Sippar texts. With this kind of data, when a model is trained only on Nippur signs, it expects two vertical wedges for an UD and thus it will have a lower prediction accuracy on UD signs from Dūr-Abiešuḫ and Sippar.

---

[12]The number of classes present in our corpus is higher than what is sometimes assumed a scribe in the Old Babylonian period would know [5]:20.
[13]A discussion of the qualities of the data will be offered by Smidt forthcoming.

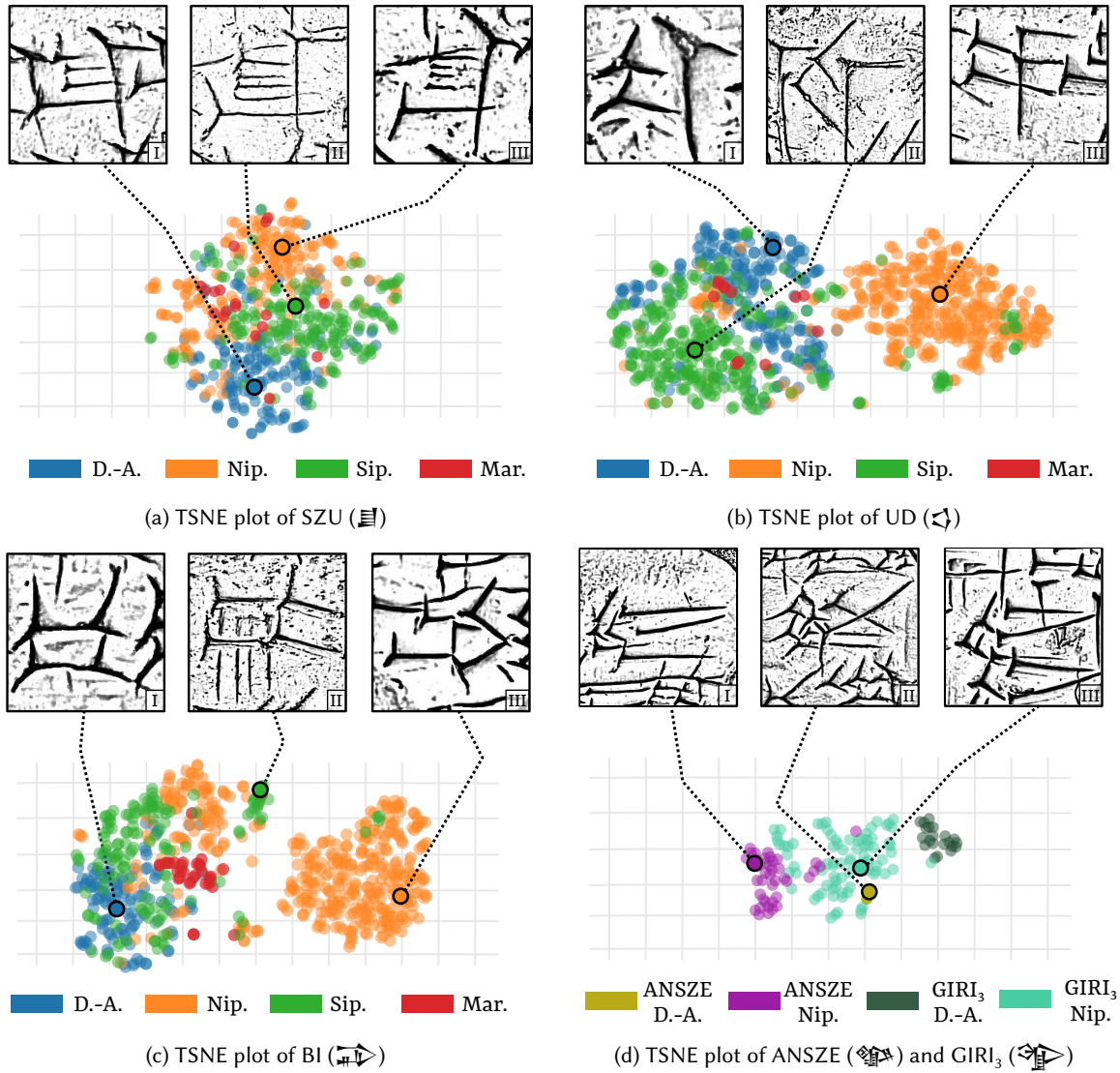

Fig. 6. TSNE visualization of the representations of a model trained on the SketchB visualizations of the three datasets together. For some signs all feature representations form a cluster, other signs are clustered into more than one cluster.

A third example is the clustering of the sign BI (𒁉), which has two dominant parameters changing across its clustered variations. In Figure 6c, I&II, the right part of the sign consists of two corner wedges with Figure 6c, III consisting of two wedges directed towards each other. As for UD is it for BI, the Nippur attestations dominates one of the clusters with a few Sippar attestations interspersing the cluster. The left cluster i.e., Figure 6c, I&II, with two corner wedges is especially differing on the number of smaller verticals inside the sign and the cursiveness. Again, the variations are similar to those found in Labat [17]:122. The classification of ANSZE (𒀲) is not always straight-forward for our model as can be seen in Figure 1c. Of the three examples, two are classified well, but the last is not suggesting ANSZE as a possibility, in fact GIRI$_3$ (𒄊) was deemed the most likely result. We therefore examined the feature representation of ANSZE and GIRI$_3$ in Figure 6d and visualized how they relate to each other. With some exceptions the ANSZE from the Nippur corpus cluster nicely together on the left and similarly does GIRI$_3$ from Dūr-Abiešuḫ on the right. However, the GIRI$_3$ from Nippur are to some degree clustering in the middle, with a substantial amount of attestations clustering closely to the ANSZE from Nippur. The ANSZE signs from Dūr-Abiešuḫ are clustered isolated on the edge of the GIRI$_3$ cluster from Nippur. We can assume that the right most example in the in Figure 1c is from the Dūr-Abiešuḫ corpus and after checking, that is indeed the case. Variations as presented by [17]:118,198 does support a possible confusion between ANSZE and GIRI$_3$. We also see that the variations can be heavily provenience dependent and that the very low number of attested ANSZE signs from Dūr-Abiešuḫ likely influenced this confusion. If we fine-tuned on the Dūr-Abiešuḫ data alone, the differences between ANSZE and GIRI$_3$ in that corpus would also allow for better classifications as within that corpus they are distinct.

The presented analyses of clusters (see Figure 6), lead us to assume that the model is reasonably good at representing sign variations similar to human intuition, instead of clustering variations based on background and image noise. With the confusion of the ANSZE signs from Dūr-Abiešuḫ, it is also showcased how important a minimum amount of attestations is or that fine-tuning to a provenience could help the model distinguish between palaeographic overlaps across sign classes, which we further demonstrate in the next section.

*4.3.3 Results by provenience.* Cuneiform tablets come from many different proveniences and writing styles may vary greatly among them. This can negatively impact the performance of an ML-model trained on one type of tablet when evaluated on another. Therefore, we discuss the transferability of models trained on a dataset of a single provenience and the improvements when more proveniences are added to the training set. One provenience, Marad, is kept outside of training at all times to give an estimate of the out-of-distribution (OOD) performance.

Figure 7 shows the results of this experiment. When training on a single provenience, performance on OOD sets is low. OOD accuracies range from 51.4% of the in-distribution performance (Sippar set after training on Dūr-Abiešuḫ data) to 79.0% (Marad set after training on Nippur data). When more diverse training sets are used (Nippur, Dūr-Abiešuḫ and Sippar sets), OOD performance increase to 93.4% of the average in-distribution performance. The (dis)similarity of the Marad tablets and the trained sets, is assumed to be roughly equal to the (dis)similarity between the different trained sets. If a less similar OOD set would be used e.g., from the south of Mesopotamia (closer to the Persian Gulf), performance may still be comparatively low.

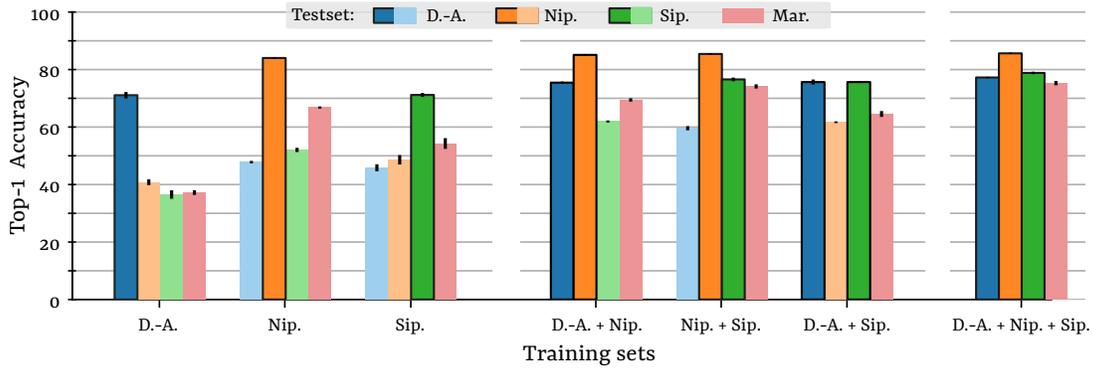

Fig. 7. Training results for different dataset combinations, split by provenience on the SketchB visualizations. Dark bars show the results of the trained proveniences, faded bars are out of distribution sets. Small black lines indicate the standard deviation of the results.

Table 1. Top-1 and top-5 accuracy of models trained on the SketchB data. The base model is trained on all three proveniences with data augmentations, this corresponds to the right-most model in Figure 7. The fine-tuned models are trained with a lower learning rate and no augmentations, which further improves the results by aligning the training data and test data better.

|  | Top-1 accuracy | | | Top-5 accuracy | | |
| --- | --- | --- | --- | --- | --- | --- |
|  | Dūr-Abiešuḫ | Nippur | Sippar | Dūr-Abiešuḫ | Nippur | Sippar |
| Base | 77.2 ± 0.2 | 85.6 ± 0.2 | 78.8 ± 0.4 | 93.6 ± 0.5 | 96.1 ± 0.1 | 91.9 ± 0.6 |
| Fine-tune all | 78.7 ± 0.4 | 86.8 ± 0.2 | 80.1 ± 0.4 | 94.0 ± 0.2 | 96.5 ± 0.1 | 92.6 ± 0.5 |
| Fine-tune Dūr-Abiešuḫ | 79.3 ± 0.4 | | | 94.2 ± 0.4 | | |
| Fine-tune Nippur | | 87.1 ± 0.3 | | | 96.5 ± 0.1 | |
| Fine-tune Sippar | | | 80.8 ± 0.4 | | | 92.2 ± 0.6 |

As accounted for in Section 3.1.2, the scholarly assumption is that the Nippur and Dūr-Abiešuḫ datasets are relatively close to each other in scribal traditions. Figure 7 indicates that training on the Nippur dataset is resulting in a better performance for the Marad texts than for the Dūr-Abiešuḫ texts. Whereas, there is only a negligible difference in performance between the Nippur and Marad sets when training on the Dūr-Abiešuḫ set. The location of Dūr-Abiešuḫ is unknown, but Nippur and Marad are only c. 40 km from each other. However, the observed differences in testing performances could be an artifact of data types, chronological spread and relative dataset sizes i.e., the Nippur set is relatively large compared to the Dūr-Abiešuḫ and Sippar set (see Figure 2). To better understand the differences between the corpora, more granularity of genre and chronological data would be necessary. However, this granularity is difficult to achieve as not all tablets are dated and as the genres are not well defined.

The previous discussion in Section 4.3.2 shows that signs can be written differently across different proveniences. This suggests that model specification, where we start from a base model trained on varied data and add a final training stage with data only specific to the provenience of interest, may lead to further improvements of these specialized models. The results of this experiment are presented in Table 1. The base model is trained on all three proveniences (Nippur, Dūr-Abiešuḫ and Sippar). This base model is then fine-tuned using either all data or only the data of one specific provenience. Crucially, this is done with a lower learning rate and without the rotation and perspective augmentations (see details in Section 3.3). The specialization step brings the training data distribution closer to the test distribution, further improving the results. For all three proveniences, the best results are attained with a final training step on only the data of that specific provenience.

## 5 Conclusions and Perspectives

### 5.1 Conclusions

With our 2D+ dataset we were able to assess the impact of various lighting angles and different depth visualisations on a cuneiform signs classification task. When light hits a sign at an angle close to perpendicular, few shadows are cast, and signs become hard to classify. Especially, for the Color00 visualisations, which with an even illumination has few clear shadows, the classification of signs is most difficult (although 74.1% top-1 accuracy is still a decent result). We observe similar issues with ColorC and ColorG, which have light directly from the right and left, respectively. The right and left of the front and back are where tablets are the most curved, and as light falling on curved parts will be closer to a perpendicular angle, the shadows are less clear, this tendency can also be seen on the normal maps. Visualizations with light coming from the top or bottom are the top scoring photorealistic visualisations, as their top-1 accuracy perform 3.5 percentage points better than Color00 and 2.1 percentage points better than ColorG i.e., light directly from the right. Comparatively, explicitly visualising depth information leads to a 4 (normal maps) and 5 (sketches) percentage points top-1 accuracy improvement over any photorealistic visualisation. Based on our results we can conclude that having at least two images with substantial different light sources available, is ideal. That light should come from the top or the bottom. Ultimately, combined light information can then be combined to create depth maps, which as mentioned, lead to a superior performance.

Our testing dataset can be divided into four subgroups defined by their proveniences. We evaluated how well a trained model generalizes on data from proveniences that were not seen during training. When training on a single provenience, the model tends to generalize poorly to data from unseen proveniences. By including more proveniences in the training data, up to three in our case, performance on unseen proveniences nearly matches those present in the training data. With the help of TSNE plots, we were able to visualise the variations within sign categories, thereby showing clear distinctions between variants. These plots indicated that our ML model is able to differ between sign variations akin to human analysis. When gathering more data, it is important to focus on the variety, and not only on the quantity of data. With more variations and more sign categories, our base model could be further improved. Finally, a more general base model can subsequently be fine-tuned with data from a specific provenience, offering superior performance for that given provenience.

### 5.2 Perspectives

The dataset presented here is of particularly high quality when it comes to visualisation types and annotation granularity. Only a limited number of cuneiform tablets have been imaged as 2D+ or 3D objects, the vast majority of legacy image data are photographs and flatbed scans. With thousands of cuneiform tablets still to be imaged, efficiency and feasibility will be of high priority for collection holders and archaeologist. Furthermore, when applying more cumbersome imaging strategies it is common to focus on tablets that are fairly nice and whole, despite a considerable part of the cuneiform corpus consisting of fragmentary objects. An important future step for sign classification research is to quantify any model's performance on flatbed scans and photographs. The logical step thereafter is to identify methodologies to transfer sign classification learning from high granularity visualisation types to the legacy data. We also hope that the introduction of our data encourages acquisition of images as 2D+ files or normal maps. As of now, it takes many calibrated images to make accurate normal maps, but technical developments, both software and hardware, could make it possible to generate them from fewer images from simpler camera setups. Additionally, further research in the use of visualisations for classification tasks could identify new types of visualisations that preserve or improve performance compared to sketches, while also making acquisition easier.

To some extent the performance of the model is still lacking for top-1 predictions, but our top-5 predictions performance for cuneiform signs is good. To further improve the quality of the model trained, our results indicate that including data from more various proveniences has a positive effect on transfer learning on tablets from yet unseen sites. A corpus with a larger geographical distribution would also allow for broader examinations of palaeographic variations with methods such as TSNE plots. Variation in the data will also likely limit the number of signs with less than 20 instances, thereby reducing the blind spots of our trained classification model. Following that, a better and more robust model could allow for few-shot classification of signs with less than 20 instances.

The classification of signs is a necessary part of a full OCR pipeline, but does not complete it. Firstly, it requires the localization of signs on a tablet, as discussed in Section 2. Secondly, the disambiguation of sign classes often require understanding of the context. Many signs have multiple possible readings, and the preceding and following signs can often help in disambiguation. Even including simple language models e.g., N-grams, may dramatically improve the top-1 classification accuracy [24]. As part of a full OCR pipeline, we are working on a suggestion tool that after the user defines a bounding box, gives a top-5 prediction of the sign (see Appendix A). In the future, our goal is to extend this tool with localisation of signs and an improved classification model that includes linguistic information.

Cuneiform OCR shares a similar issue to many OCR efforts, that numbers are usually written with numerals and therefore they are difficult to classify. Depending on what is counted, OB Akkadian texts used various number systems (see [15]:579-585), which further complicates the task of identifying them. Currently, we deemed it feasible to divide the numerals into their classes e.g., the ones by themselves, the tens by themselves and the sixties by themselves. However, many of the numbers in our dataset are attested less than 20 times, meaning they are not included in the training and in-domain testing data. Going forward we either need to curate enough new data to cover more number combinations or find a solution that does not necessitate a minimum of 20 instances of all numbers.

Lastly, implementing ML methodologies on the cuneiform data is not done in an effort to make cuneiform researchers obsolete. It is doubtful that within the current methodologies and data size it will be possible to have automatic machine reading capabilities. More likely, ML will facilitate research of cuneiform tablets by helping with difficult to read signs, by quantifying linguistic features, and by facilitating palaeographic analyses.

Acknowledgments

To all the people writing Cuneiform 4000 years ago. We thank Belspo for funding our project and essentially allowing this work to be done. A thanks goes to the Royal Museums of Art and History (RMAH) in Brussels, partner in our project and provider of access to the Marad testing data. Our appreciation also goes out to Olivier Goossens (KU Leuven) for preparing the datasets of visualisations of all the cuneiform tablets based on 2D+ images, Anne Goddeeris (Ghent University) for being part of the annotation crew, and Timo Homburg (Mainz University of Applied Sciences) for providing access to and support for the Cuneur annotation software. Frederic Lamsens (Ghent Centre for Digital Humanities) have made the visualisations available for annotations. The collections that allowed for 2D+ images to be acquired were crucial: Vorderasiatische Museum in Berlin, the Department of Near Eastern Studies at Cornell University and the Frau Professor Hilprecht Collection of Babylonian Antiquities at the Friedrich-Schiller University in Jena.

## A  Cune-AI-form Tool

As part of the dissemination of this work, we release a first version of a Cune-AI-form classification tool. A screenshot of the tool is shown in Figure 8. The user can use this tool to localize a sign in the tablet, for which our trained machine learning model will then predict the sign that is visible within the selected rectangle. This tool can be run locally as a python script, for which all the code is made publicly available here. The user can then either use the tablets that were used as validation data in this paper, or use their own tablets. At its current stage, this tool can already be useful in classifying signs from proveniences a researcher or other type of users are not familiar with. In the future we plan to further extend this tool by incorporating aids for localizing the signs, as well as taken preceding and following signs into account for the prediction.

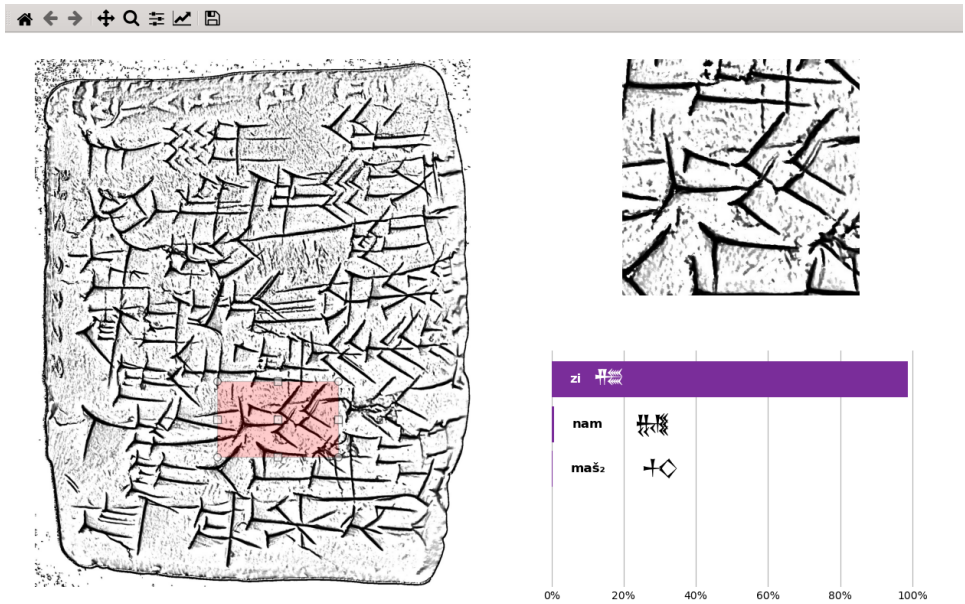

Fig. 8. Screenshot from the first version of the Cune-AI-form tool. The user draws the (red) rectangle on the tablet on a sign they would like to classify. The crop is then shown in the top right of the tool and the predictions made by the model are on the bottom right. Producing the predictions for one crop take less than a second on a modern laptop, without dedicated GPU.